\documentclass[letterpaper,journal]{IEEEtran}
\usepackage{amsmath,amsfonts}
\usepackage{algorithmic}
\usepackage{algorithm}
\usepackage{array}
\usepackage[caption=false,font=normalsize,labelfont=sf,textfont=sf]{subfig}
\usepackage{textcomp}
\usepackage{stfloats}
\usepackage{url}
\usepackage{verbatim}
\usepackage{graphicx}
\usepackage{cite}
\usepackage{hyperref}
\begin{document}

\title{Semantic-Aware Generative Image Transmission\\ for Resource-Constrained Visual IoT Systems}

\author{ Chenyang Zhang, Changwang Liu ~\IEEEmembership{Member,~IEEE}, Jinqi Zhu, Jiayi Chang, Yuxuan Wang, Shuqing He and  Jia Guo* ~\IEEEmembership{Member,~IEEE} 

 \thanks{This research work was substantially supported by  the National Natural Science Foundation of China under Grant No. 62002263, Tianjin Science and Technology Program Projects: 24YDTPJC00630.}
 \thanks{ C. Zhang, C. Liu, J. Zhu  J. Chang, Y. Wang and J. Guo,are with the School of Computer and Information Engineering, Tianjin Normal University, Tianjin, China (e-mail: 2411090038@stu.tjnu.edu.cn, 004967@tjnu.edu.cn, jsjzhujinqi@tjnu.edu.cn,2511090041@stu.tjnu.edu.cn, 2511090047@stu.tjnu.edu.cn  and c04s316@bupt.cn).}
 \thanks{S. He is School of Information Science and Engineering, Linyi University, Linyi 276000, China (email:heshuqing@lyu.edu.cn)}
 \thanks{*Corresponding Author: Jia Guo} 
 \thanks{Our code is available at link: \url{https://github.com/zzzccy1/Semantic-Aware-Generative-Image-Transmission-for-Resource-Constrained-Visual-IoT-Systems.git}}
}

% The paper headers
\markboth{IEEE Internet of Things Journal}%
{Zhang: Semantic-Aware Generative Image Transmission for Visual IoT Systems}

\maketitle

\begin{abstract}
Resource-constrained visual Internet of Things (IoT) systems, such as edge cameras, unmanned sensing platforms, industrial inspection nodes, and remote monitoring sensors, often need to transmit task-relevant visual evidence over low-rate wireless links to an edge/cloud service. Existing image communication methods usually compress or transmit complete global representations, leaving limited room to exploit receiver-side generative restoration. This paper proposes a semantic-aware generative image transmission framework for edge-assisted visual IoT. The image captured by an IoT visual sensor is encoded into a discrete token grid by a VQ encoder. At the IoT transmitter or nearby gateway, token recoverability, estimated from prediction entropy and local structure complexity, is fused with semantic importance obtained from instance segmentation and category-aware scoring. A spatial dispersal sampler then selects the tokens to be transmitted under a bitrate budget. The transmitter sends only the quantization indices of kept tokens and a binary mask map, while the edge/cloud receiver recovers masked tokens through MaskGIT with Halton sequence scheduling. Experiments on Kodak and VisDrone scenes under AWGN and Rayleigh channels show that the proposed method provides a flexible bitrate--quality tradeoff for narrowband visual IoT links. At 0.074 bpp, it uses 44.6\% of the transmitted bits of the 0.167-bpp DeepJSCC/WITT reference while achieving 29.9 dB PSNR. A pseudo-GT downstream detection study on Kodak further shows that semantic-aware masking preserves task-relevant objects better than random masking at both 30\% and 50\% mask ratios.
\end{abstract}


\section{Introduction}

Visual data---images and video---are becoming a dominant sensing modality in the Internet of Things (IoT), driven by smart cameras, unmanned aerial platforms, industrial inspection nodes, remote environmental monitoring, and edge visual analytics. A typical visual IoT service consists of a sensing node, a bandwidth- and energy-constrained wireless uplink, and an edge gateway or cloud service that performs monitoring, inspection, or decision support. In such systems, the receiver often cares about object identity, scene semantics, and task-relevant evidence rather than pixel-perfect reconstruction. Conventional communication systems follow Shannon's separation principle~\cite{Shannon48} and aim to minimize bit errors under channel capacity constraints. However, faithfully transmitting complete visual representations can carry substantial redundancy for IoT perception tasks and makes it difficult to jointly satisfy low latency, low bandwidth, and communication-side energy constraints. The central challenge is therefore: how can a visual IoT device reduce unnecessary visual payload while preserving the receiver's ability to interpret critical semantic information?

Semantic communication addresses this challenge by shifting the objective from bit accuracy to semantic fidelity. Instead of transmitting every bit faithfully, it prioritizes task-relevant information and reduces redundant overhead, which is especially attractive for battery-powered or low-rate visual IoT links. For visual IoT image transmission, the key question becomes: which parts of an image carry essential semantics for downstream monitoring or decision making, and which parts can be omitted or coarsely described? In complex natural scenes, foreground objects such as persons, vehicles, animals, and industrial targets often carry disproportionate semantic weight relative to background textures. A visual IoT transmitter that performs fine-grained semantic screening---coupled with an edge/cloud receiver that restores missing content---can reduce wireless payload without sacrificing semantic fidelity. Image semantic communication for visual IoT is therefore a joint optimization of \textit{information selection} and \textit{semantic preservation}: maximizing the semantic value delivered to the edge receiver per unit of wireless bandwidth.

Generative models~\cite{Huynh24} offer a concrete mechanism for realizing this vision in edge-assisted visual IoT. Discrete visual representations---VQ-VAE~\cite{VanDenOord17}, VQGAN~\cite{Esser21}---encode images as compact token sequences, making visual content amenable to token-level manipulation. MaskGIT~\cite{Chang22}, a masked token prediction model, recovers masked tokens through parallel decoding, offering faster inference than autoregressive alternatives while maintaining high reconstruction quality. This suggests a natural IoT-edge architecture: an IoT visual sensor or nearby gateway selectively transmits a subset of tokens, and the edge/cloud receiver reconstructs the missing ones using MaskGIT. The approach overcomes a fundamental limitation of conventional image transmission---that all information must be explicitly sent over the wireless link.

However, MaskGIT was designed for image generation, where masks are applied randomly and uniformly to train the model's generative capability. This random masking strategy is ill-suited for visual IoT communication for three reasons. First, it ignores recovery difficulty: tokens in textured, high-frequency regions are harder to reconstruct than those in smooth areas, yet random masking treats them identically. Second, it offers no semantic protection: foreground objects relevant to surveillance, inspection, or monitoring can be masked as readily as background, risking the loss of critical IoT sensing content. Third, it neglects spatial distribution: clustered masks destroy local context, degrading reconstruction stability at the edge/cloud receiver. The core problem is that MaskGIT's masking mechanism answers \textit{how to generate}, not \textit{which tokens deserve scarce IoT transmission bandwidth}. Meanwhile, reconstruction metrics such as PSNR alone are insufficient to evaluate visual IoT communication quality, since an image that appears visually faithful may still suffer semantic distortion in foreground regions. An intelligent token selection mechanism that combines generative reconstruction capability with semantic resource allocation principles is needed.

We propose a semantic-aware generative image transmission method for resource-constrained visual IoT systems. The method operates on discrete token grids produced by a VQ encoder and jointly models two complementary signals: \textit{token recoverability}, estimated via prediction entropy and local structure complexity, and \textit{semantic importance}, derived from instance segmentation and category-aware scoring. These two signals are fused and combined with a spatial dispersal sampling procedure to produce the final mask. The IoT transmitter or edge gateway then sends only the content of kept tokens and a binary mask map; the edge/cloud receiver recovers masked tokens using MaskGIT with Halton sequence scheduling. By protecting semantically important and hard-to-recover tokens while offloading redundant, easily-predicted content to the edge/cloud-side generative model, the method achieves a favorable trade-off between wireless bandwidth consumption and semantic reconstruction quality.

The main contributions of this paper are as follows:

\begin{enumerate}
    \item \textbf{A semantic-aware token selection method for visual IoT transmission.} We jointly model token recovery difficulty (via prediction entropy and local structure complexity) and semantic importance (via instance segmentation and category-aware scoring), converting the conventional random masking strategy into an intelligent token selection mechanism for bandwidth-constrained IoT links. Score fusion and spatial dispersal sampling allocate scarce transmission bandwidth preferentially to semantically critical and hard-to-recover regions, while redundant, easily-predicted content is completed by the edge/cloud generative model.

    \item \textbf{An edge-assisted generative image communication framework for resource-constrained visual IoT.} Through a ``key tokens transmitted, non-key tokens generated'' paradigm, the IoT transmitter sends only the quantization indices of kept tokens and a binary mask map; the edge/cloud receiver recovers the full image via MaskGIT with Halton sequence scheduling. This decoupled architecture separates \textit{what to transmit} (semantic token selection at the IoT side or gateway) from \textit{how to reconstruct} (generative restoration at the edge/cloud side), supporting flexible adaptation to SNR conditions, mask ratios, and visual IoT scene categories.

    \item \textbf{A communication-aware evaluation protocol for visual IoT links.} We include mask-map overhead in the transmitted payload, evaluate AWGN and Rayleigh fading channels, report communication-side delay and RF-energy proxies, and further measure downstream object detection accuracy. These experiments are intended to assess not only pixel fidelity but also task-relevant semantic preservation under constrained IoT bandwidth.
\end{enumerate}

\section{Related Work}

\subsection{Semantic Communication for Visual IoT}

Semantic communication shifts the objective from bit-level accuracy to semantic fidelity, prioritizing task-relevant information under limited channel resources~\cite{Getu24,Nguyen25}. It is particularly relevant to visual IoT, where cameras and inspection sensors must deliver semantic evidence over low-rate or energy-constrained links~\cite{Wang25,Zhong25}. Recent visual semantic communication frameworks jointly optimize encoding, channel transmission, and edge/cloud reconstruction or downstream tasks~\cite{Zhao25}, using CNN/Transformer encoders, channel-adaptive modulation, or task-oriented losses.

However, most existing methods emphasize global feature compression and still transmit complete feature representations, leaving limited ability to exploit edge/cloud-side generative restoration. Although GAN- and diffusion-assisted schemes~\cite{Liang24,SING25,Yuan25} show that generative priors can compensate for missing content, token-level semantic selection for resource-constrained visual IoT remains insufficiently explored.

\subsection{Joint Source-Channel Coding}

Joint Source-Channel Coding (JSCC) directly maps image pixels to channel symbols without explicit source and channel codecs. DeepJSCC~\cite{Bourtsoulatze19} introduced CNN-based wireless image transmission, while WITT~\cite{Yang23}, SwinJSCC~\cite{SwinJSCC24}, Mamba-based coding~\cite{MNTSCC25}, and cooperative relay schemes~\cite{Bian24} further improved learned transmission. Learned compression and JSCC surveys also show rapid progress in this direction~\cite{Huang24,Gunduz24}. Nevertheless, JSCC methods usually send a complete set of channel symbols and do not explicitly allocate bandwidth according to spatial semantic importance, motivating hybrid selective-transmission and generative-restoration designs.

\subsection{Generative Visual Reconstruction}

Discrete visual representations bridge image generation and efficient transmission. VQ-VAE~\cite{VanDenOord17} and VQGAN~\cite{Esser21} quantize images into compact codebook indices, and recent tokenizers~\cite{TokenFlow25,UniTok25,MaskBit25} further improve discrete representation learning. Such token grids (e.g., $24 \times 24$ for a $384 \times 384$ input) naturally support token-level manipulation and selective transmission~\cite{Li25TokenSurvey}.

MaskGIT~\cite{Chang22} recovers masked tokens through parallel bidirectional Transformer decoding~\cite{Vaswani17}. Unlike autoregressive models, it updates masked positions in parallel and can restore a partially transmitted token grid within a fixed number of decoding steps, making it attractive for edge/cloud-side reconstruction.

However, MaskGIT was designed with random masks for generation. In visual IoT communication, masks also determine link resource allocation: textured or contour regions are harder to recover, and foreground objects carry more task-relevant information than background. Therefore, the communication setting requires mask design that jointly considers recoverability, semantic importance, and spatial coverage.

\section{Proposed Framework}

\begin{figure*}[t!]
    \centering
    \includegraphics[width=1.0\textwidth]{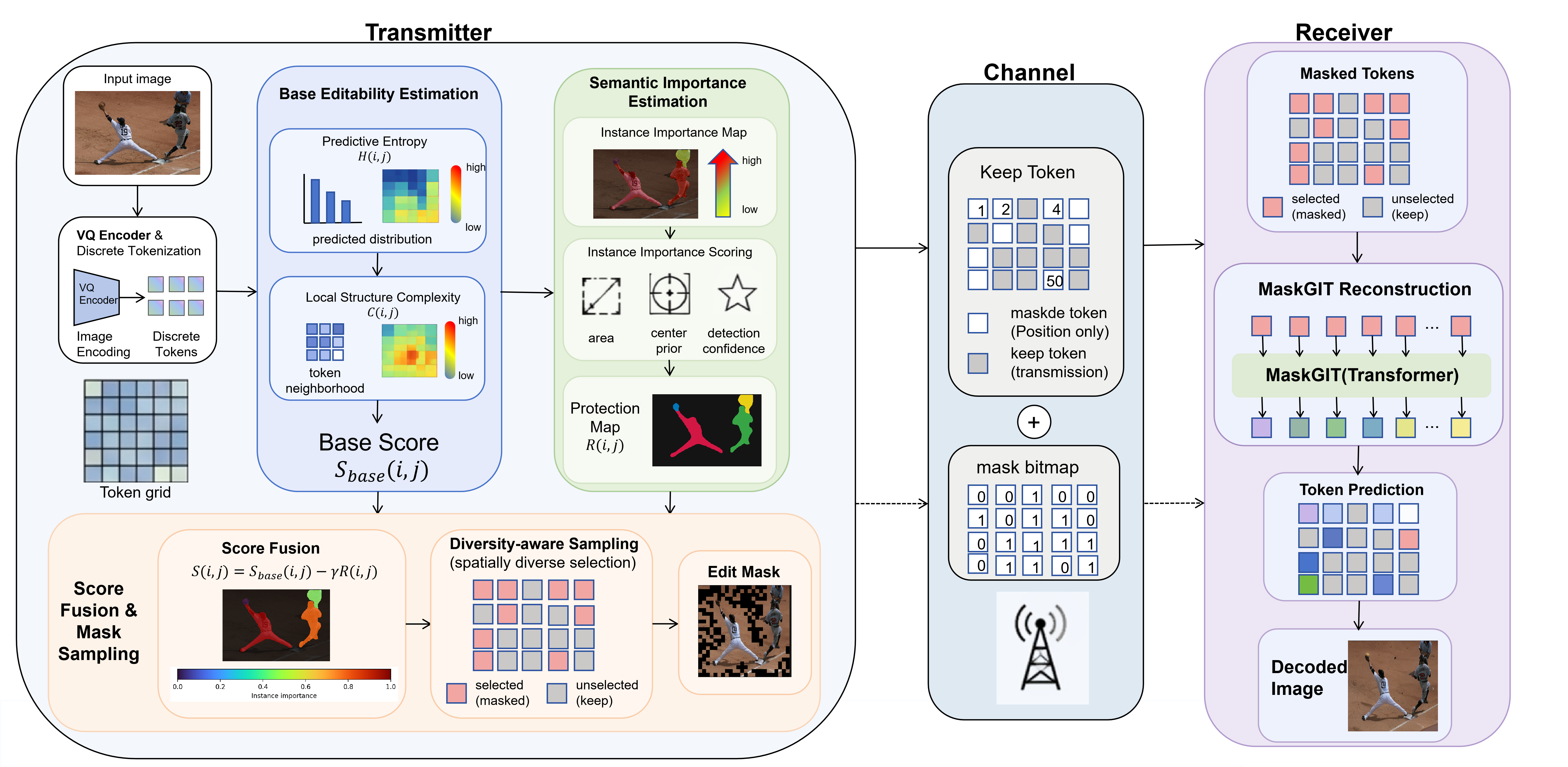}
    \caption{Overall architecture of the proposed semantic-aware generative image transmission framework for resource-constrained visual IoT systems.}
    \label{fig:framework}
\end{figure*}

The proposed method targets an edge-assisted visual IoT scenario in which a sensing node or a nearby gateway captures an image, transmits selected visual tokens over a constrained wireless uplink, and relies on an edge/cloud receiver for generative restoration. This deployment model is consistent with practical visual IoT systems: the battery-powered camera may perform only lightweight acquisition and token packaging, while a gateway or edge server can execute heavier semantic analysis and reconstruction when available. The communication payload, however, is always counted on the wireless IoT link between the sensing side and the receiver side. The framework operates in three stages. First, the input image is encoded into a discrete token grid via a VQ encoder. Second, at the IoT transmitter or edge gateway, base editability estimation, semantic importance protection, score fusion, and spatial dispersal sampling collaboratively produce a binary mask map, and the transmitter sends only the quantization indices of kept tokens together with the mask map. Third, the edge/cloud receiver recovers masked tokens via MaskGIT with Halton sequence scheduling and reconstructs the image through the VQ decoder. Fig.~\ref{fig:framework} illustrates the overall architecture.

Let the input image be
\begin{equation}
X \in \mathbb{R}^{H\times W\times 3}
\label{eq4-1}
\end{equation}

The VQ encoder first maps the image to a discrete token grid:
\begin{equation}
Z=\mathrm{VQEncoder}(X), \qquad Z\in \mathbb{Z}^{H_t\times W_t}
\label{eq4-2}
\end{equation}

where $H_t$ and $W_t$ denote the height and width of the token grid, respectively. The system sequentially performs base editability estimation, semantic importance protection, score fusion, and spatial dispersal sampling at the IoT transmitter or edge gateway to obtain the final mask set. For kept tokens, their content indices are transmitted; for masked tokens, the binary mask map indicates their positions without sending their token values. The edge/cloud receiver constructs a masked token grid from the received information, feeds it into the MaskGIT model to predict the missing tokens, and finally recovers the image through the decoder.

\subsection{Image Encoding and Discrete Token Representation}

In conventional image transmission, directly treating pixels as communication objects often incurs substantial data redundancy, which is unfavorable for efficient transmission in resource-constrained scenarios. To this end, this paper first converts the input image into a discrete token representation, mapping the image content from the continuous pixel space to a compact semantic representation space.

Specifically, the input image passes through an encoder to obtain continuous latent variables, which are then discretized via a codebook. For position $(i,j)$ in the token grid, the corresponding discrete token can be expressed as
\begin{equation}
z_{i,j}=\arg\min_{e_k\in \mathcal{C}} \|E(X)_{i,j}-e_k\|_2^2
\label{eq4-3}
\end{equation}

where $\mathcal{C}=\{e_k\}_{k=1}^{K}$ denotes the codebook, $K$ is the codebook size, and $E(X)_{i,j}$ represents the continuous feature output by the encoder at position $(i,j)$. Through this process, the entire image is represented as a two-dimensional grid of discrete tokens:
\begin{equation}
Z=\{z_{i,j}\mid 1\le i\le H_t,\ 1\le j\le W_t\}
\label{eq4-4}
\end{equation}

This representation offers two advantages. First, the token grid significantly reduces the image representation dimensionality, transforming the transmission object from pixel-level data to more compact semantic units. Second, discrete tokens are naturally compatible with generative models such as MaskGIT, enabling the receiver to complete missing information in the token space. Therefore, all subsequent mask decisions and transmission operations in this paper are performed on the token grid $Z$.

\subsection{Semantic-Aware Mask Generation}
\label{subsec:mask_gen}

\textbf{1) Base Editability Estimation.} In visual IoT semantic communication scenarios, not all tokens are equally suitable for masking. For tokens with clear contextual relationships and simple local structures, even if the IoT transmitter does not send their specific content, the edge/cloud receiver can usually recover them fairly accurately using the generative model. In contrast, for edge details, texturally complex regions, or object contour areas, direct masking can easily cause cumulative recovery errors. Therefore, before generating the final mask, it is necessary to evaluate the base recoverability of each token.

This paper refers to this process as base editability estimation. The module characterizes the recovery difficulty of a token from two aspects: MaskGIT's prediction uncertainty for that token, and the structural complexity of the local region where the token resides. Combining both, the base editability score for the token at position $(i,j)$ is defined as
\begin{equation}
S_{\text{base}}(i,j)=\alpha H(i,j)+\beta C(i,j)
\label{eq4-5}
\end{equation}

where $H(i,j)$ denotes the prediction entropy, $C(i,j)$ denotes the local structure complexity, and $\alpha$ and $\beta$ are weighting coefficients.

Prediction entropy quantifies the uncertainty of MaskGIT's prediction for a token at a given position under contextual conditions. If the prediction distribution at a position is relatively concentrated, it indicates that the token can be more easily recovered from neighborhood information; conversely, if the prediction distribution is dispersed, the recovery risk is higher. Correspondingly, the prediction entropy can be written as
\begin{equation}
H(i,j)= -\sum_{k=1}^{K} p(k|\mathcal{N}_{i,j})\log p(k|\mathcal{N}_{i,j})
\label{eq4-6}
\end{equation}

where $\mathcal{N}_{i,j}$ denotes the contextual neighborhood of position $(i,j)$, and $p(k|\mathcal{N}_{i,j})$ is the predicted probability of codebook index $k$ at position $(i,j)$ given $\mathcal{N}_{i,j}$.

Beyond prediction uncertainty, local structure complexity also affects restoration quality. For smooth background regions, even if a token is masked, the model can relatively easily complete it using surrounding context; however, at locations with edges, textures, and pronounced local structural variations, the absence of a single token tends to have a greater impact on visual continuity. Therefore, this paper further characterizes local complexity using token differences within the neighborhood:
\begin{equation}
C(i,j)=\frac{1}{|\mathcal{N}_{i,j}|}\sum_{(u,v)\in \mathcal{N}_{i,j}} d(z_{i,j},z_{u,v})
\label{eq4-7}
\end{equation}

where $d(\cdot,\cdot)$ denotes a difference metric between tokens.

Overall, a larger base editability score $S_{\text{base}}(i,j)$ indicates that the token is more difficult to recover from context and is thus more suitable for direct transmission as a keep token; conversely, a smaller score suggests it is more suitable as a masked token to be left for edge/cloud-side prediction. It should be noted that this module primarily reflects ``recovery risk'' and does not yet explicitly consider semantic-level importance; therefore, a semantic protection mechanism needs to be further introduced.

\textbf{2) Semantic Importance Protection.} The goal of semantic communication is not merely to minimize reconstruction error, but more importantly to ensure that the key foreground objects, core targets, and primary scene information in the image are not corrupted. If only base editability estimation is relied upon, the system may focus excessively on ``which regions are easier or harder to recover'' while neglecting ``which regions are semantically more important.'' Therefore, this paper further designs a semantic importance protection mechanism to preferentially preserve critical object regions in the image.

First, instance segmentation is performed on the input image, yielding a set of objects
\begin{equation}
\mathcal{O}=\{o_n\}_{n=1}^{N}
\label{eq4-8}
\end{equation}

where $N$ is the number of detected instances in the image. For each instance, this paper integrates four factors---area, centrality, detection confidence, and category prior---to construct its instance importance score:
\begin{equation}
I_n=\eta_1 A_n+\eta_2 Q_n+\eta_3 D_n+\eta_4 P_n
\label{eq4-9}
\end{equation}

where $A_n$ denotes the area score, $Q_n$ denotes the centrality score, $D_n$ denotes the detection confidence, $P_n$ denotes the category prior, and $\eta_1=\eta_2=\eta_3=\eta_4=1$ (equal weights; the four factors are normalized and summed with a uniform prior).

The area score reflects the proportion of the instance in the image, the centrality score reflects whether its position is close to the visual center, the detection confidence measures the reliability of the object recognition result, and the category prior reflects the relative importance of different semantic classes in the communication task. For instance, in most natural images, foreground objects such as persons and vehicles are assigned higher category prior weight than background regions.

To enable this semantic information to participate in token-level mask decisions, the instance importance must be further mapped onto the token grid. Accordingly, the semantic protection score at position $(i,j)$ is defined as
\begin{equation}
R(i,j)=\sum_{n=1}^{N} I_n\cdot \mathbb{I}\big((i,j)\in o_n\big)
\label{eq4-10}
\end{equation}

where $\mathbb{I}(\cdot)$ is the indicator function. When a token lies within a high-importance instance region, its $R(i,j)$ value is large, indicating that this position should be avoided from masking whenever possible.

Through this mechanism, the transmitter no longer treats all tokens equivalently but can consciously protect critical semantic regions in the image. Consequently, the subsequent mask generation considers not only recovery difficulty but also the fundamental principle of semantic communication: ``prioritizing the preservation of important information.''

\textbf{3) Score Fusion and Spatial Dispersal Sampling.} After obtaining the base editability score and the semantic protection score, this paper further fuses them to determine the final mask positions. The core idea of the fusion is that tokens with high recovery difficulty should tend to be directly transmitted, and tokens with high semantic importance should also avoid being masked. Therefore, the keep-priority score for a token is defined as
\begin{equation}
S_{\text{keep}}(i,j)=S_{\text{base}}(i,j)+\gamma R(i,j)
\label{eq4-11}
\end{equation}

where $\gamma$ is the semantic protection weight. This formulation indicates that when a position has high recovery difficulty or high semantic importance, its keep priority increases, making it less likely to be selected as a masked token. To balance the trade-off between token recoverability and semantic importance, the weight parameter $\gamma$ is introduced. When $\gamma$ is small, token selection favors editability; when $\gamma$ is large, the model tends to protect semantically critical regions. In the experiments, $\gamma = 0.5$ is used.

Given a mask ratio $\rho$, and with a total number of tokens $M=H_tW_t$, the number of masked tokens to be selected is
\begin{equation}
M_m=\lfloor \rho M \rfloor
\label{eq4-12}
\end{equation}

The most straightforward approach is to sort tokens by their keep-priority scores and mask the lowest-priority tokens. However, this can lead to a problem: the selected tokens may be overly concentrated spatially, resulting in large missing regions locally and making it difficult for the receiver to recover stable contextual relationships. To address this, the paper further introduces a spatial dispersal sampling strategy.

Specifically, when selecting masked tokens, not only is the composite score considered but also the spatial distance between candidate positions and already-selected mask positions. Positions too close to existing masked tokens have their priority reduced, while more uniformly distributed positions are preferentially retained. Accordingly, a spatial penalty term is defined as:
\begin{equation}
D(i,j)=\sum_{(u,v)\in \Omega_m^{(t)}}
\exp\left(-\frac{\|\mathbf{p}_{i,j}-\mathbf{p}_{u,v}\|_2^2}{\sigma^2}\right)
\label{eq4-13}
\end{equation}

where $\Omega_m^{(t)}$ denotes the set of currently selected masked tokens, $\mathbf{p}_{i,j}$ denotes the spatial coordinates of position $(i,j)$, and $\sigma$ is a parameter controlling the neighborhood influence range. After incorporating the spatial constraint, the iterative mask utility becomes
\begin{equation}
U(i,j)=-S_{\text{keep}}(i,j)-\mu D(i,j)
\label{eq4-14}
\end{equation}

where $\mu$ is the dispersal weight. At each iteration, the candidate with the largest $U(i,j)$ is selected as a masked token. Through this strategy, masked tokens can be more uniformly distributed across the entire image, avoiding extensive local context loss and thereby improving the reconstruction stability at the receiver.

\subsection{Semantic Communication over the Visual IoT Link}
\label{subsec:transmission}

Based on the final mask map $\mathbf{M} \in \{0,1\}^{H_t \times W_t}$ obtained from the score fusion and spatial dispersal sampling described in Section~\ref{subsec:mask_gen}, the token grid is partitioned into keep tokens ($M_i = 0$) and masked tokens ($M_i = 1$). For keep tokens, the IoT transmitter or gateway directly transmits their discrete content over the wireless link; for masked tokens, the mask map carries their positions without sending the specific token values. The IoT-side masked token grid is thus constructed as:
\begin{equation}
Z_m(i,j)=
\begin{cases}
z_{i,j}, & (i,j)\in \Omega_k \\
[\text{MASK}], & (i,j)\in \Omega_m
\end{cases}
\label{eq4-15}
\end{equation}

where $\Omega_k$ denotes the set of keep tokens, $\Omega_m$ denotes the set of masked tokens, satisfying $\Omega_k \cup \Omega_m=\Omega$ and $\Omega_k \cap \Omega_m=\emptyset$. The information actually output by the transmitter can be expressed as:
\begin{equation}
\mathcal{T}=\{\mathcal{Z}_{\text{keep}},\mathcal{P}_{\text{mask}}\}
\label{eq4-16}
\end{equation}

where $\mathcal{Z}_{\text{keep}}$ is the content set of kept tokens and $\mathcal{P}_{\text{mask}}$ is represented by the binary mask map. Specifically, the visual IoT payload consists of the following two parts.

{\bfseries (1) Quantization Indices of Kept Tokens (Content Information).} For the $N_{\text{keep}}$ kept tokens with $M_i = 0$, the transmitter directly transmits their VQ quantization indices. Each token index is drawn from a codebook of size $K = 16384$, with a quantization precision of 14 bits ($\lceil\log_2 16384\rceil = 14$). Considering that practical IoT links require additional channel coding redundancy to ensure reliable transmission, this paper conservatively estimates an equivalent overhead of 20 bits per token:
\begin{equation}
B_{\text{content}} = N_{\text{keep}} \times 20 \quad \text{(bit)}
\label{eq:b_content}
\end{equation}

{\bfseries (2) Binary Mask Map (Positional Information).} The edge/cloud receiver must know which of the 576 tokens are masked and which are kept in order to perform targeted recovery of missing tokens. For this purpose, the transmitter additionally sends a 576-bit binary mask map, with each token occupying 1 bit ($0 = \text{keep},\; 1 = \text{mask}$):
\begin{equation}
B_{\text{mask}} = 576 \quad \text{(bit)}
\label{eq:b_mask}
\end{equation}

It is worth noting that if the mask pattern exhibits spatial sparsity, the binary mask map can be further compressed via lossless compression (e.g., arithmetic coding) to approximately 300--450 bits. To maintain conservatism and simplicity in estimation, all calculations below adopt the full 576-bit mask map as the mask transmission overhead.

In summary, the total number of bits actually transmitted over the visual IoT link is:
\begin{equation}
B_{\text{total}} = B_{\text{content}} + B_{\text{mask}}
                 = N_{\text{keep}} \times 20 + 576 \quad \text{(bit)}
\label{eq:b_total}
\end{equation}

where $N_{\text{keep}} = 576 \times (1 - \rho)$, with $\rho$ being the mask ratio and the result rounded up. Normalizing the total bits to per-pixel yields the bits per pixel (bpp):
\begin{equation}
\text{bpp} = \frac{B_{\text{total}}}{H \times W}
           = \frac{N_{\text{keep}} \times 20 + 576}{147456}
\label{eq:bpp}
\end{equation}

where $H = W = 384$ is the image resolution, and $H \times W = 147456$ pixels.

For the evaluated mask ratios $\rho\in\{5\%,10\%,20\%,30\%,50\%,70\%\}$, Eq.~\eqref{eq:bpp} gives a visual IoT payload range from 0.027 bpp to 0.078 bpp. The 50\% and 70\% mask configurations operate below 0.05 bpp, which is desirable for narrowband visual IoT links. By transmitting content only for semantically critical and hard-to-recover tokens, and leaving recoverable tokens to the edge/cloud generative model, the proposed method reduces link overhead while preserving key semantic information.

\subsection{Receiver-Side Reconstruction}

Upon receiving $\mathcal{Z}_{keep}$ and $\mathcal{P}_{mask}$, the edge/cloud receiver first reconstructs the masked token grid based on the kept token content and mask position information, and feeds it as input to the MaskGIT model. Specifically, the receiver assigns known values to keep token positions and retains mask markers at masked token positions, thereby forming a discrete token grid to be restored.

Subsequently, the receiver employs MaskGIT to predict the masked tokens, yielding a complete token grid:
\begin{equation}
\hat{Z}=\mathrm{MaskGIT}(Z_m)
\label{eq4-17}
\end{equation}

where $\hat{Z}$ denotes the complete token representation after edge/cloud-side restoration. Finally, the recovered token grid is mapped back to image space via the decoder, producing the reconstructed image:
\begin{equation}
\hat{X}=\mathrm{VQDecoder}(\hat{Z})
\label{eq4-18}
\end{equation}

In this process, keep tokens provide authentic semantic anchors for the edge/cloud receiver, enabling the model to maintain strong constraints in critical regions; meanwhile, masked tokens are completed by the generative model based on spatial context and semantic associations. Since the IoT transmitter or gateway has already preferentially preserved tokens that are semantically important and difficult to recover, the receiver mainly restores regions that are relatively easy to predict, which contributes to improved overall reconstruction quality and semantic consistency.

Thus, the proposed method forms a collaborative mechanism of ``intelligent selection at the IoT side, generative restoration at the edge/cloud side'' at the system level, positioning MaskGIT not merely as a conventional image generation model but as a key reconstruction module within the visual IoT semantic communication framework.

\section{Experimental Results and Analysis}
\label{sec:experiments}

\subsection{Experimental Setup}
\label{subsec:setup}

\subsubsection{Dataset and Preprocessing}
Kodak is used as the main test set, covering 24 diverse images commonly used in image compression and communication. VisDrone2019-DET test-dev is additionally used to evaluate cross-domain reconstruction on drone-captured IoT scenes. All images are resized to $384 \times 384$ and quantized into $24 \times 24 = 576$ tokens from a codebook of size $K = 16384$. No training or fine-tuning is performed on Kodak or VisDrone.

\subsubsection{Channel Model and SNR Settings}
The channel model adopts AWGN with QPSK modulation (2 bits per channel symbol), representing a controlled abstraction of a narrowband visual IoT uplink. The test SNR is set to
\begin{equation}
\mathrm{SNR} \in \{1, 4, 7, 10, 13, 15\}\ \mathrm{dB},
\end{equation}
with 13 dB as the baseline. A Rayleigh block-fading channel $y=hx+n$ is also tested, where $h \sim \mathcal{CN}(0,1)$ is constant within each coherence block and MMSE equalization is applied. All comparison and ablation experiments use identical channel settings.

\subsubsection{Mask Ratios and Bitrate Configurations}

This paper evaluates six mask ratios, $\rho \in \{5\%, 10\%, 20\%, 30\%, 50\%, 70\%\}$, corresponding via Eq.~\eqref{eq:bpp} to approximately 0.027--0.078 bpp. The channel robustness experiments additionally include the unmasked 0\% configuration at approximately 0.082 bpp for equal-bitrate comparison with JSCC baselines, while the rate-distortion experiments vary $\rho$ to obtain multiple bitrate points.

\subsubsection{Implementation Details}

Inference is performed on a single NVIDIA RTX 4090 to emulate gateway/edge-assisted deployment. The VQ encoder uses a pre-trained VQGAN with a $24 \times 24$ token grid and codebook size 16384. Semantic masks combine entropy/structure-based editability with Detic-based semantic heatmaps; Detic is used only for mask generation. Downstream detection uses a separate Mask R-CNN evaluator to avoid evaluator leakage. MaskGIT uses Halton scheduling~\cite{Halton60,Besnier25}, 32 sampling steps, and classifier-free guidance weight $w=2.0$. Energy and latency tables report communication-side costs only.

\subsubsection{Evaluation Metrics}

The evaluation uses:
\begin{itemize}
    \item \textbf{Image quality}: PSNR, SSIM, and LPIPS~\cite{HernandezCamara25}.
    \item \textbf{Bitrate}: bits per pixel (bpp), computed via Eq.~\eqref{eq:bpp}.
    \item \textbf{Communication efficiency metrics}: Communication delay $T_{\text{comm}}$ (ms) and communication-side relative energy consumption $E_{\text{comm}}$ (\%), defined respectively as
    \begin{equation}
    T_{\text{comm}} = \frac{B_{\text{total}}}{R_{\text{bw}}},\quad
    E_{\text{comm}} = \frac{B_{\text{total}}}{B_{\text{ref}}} \times 100\%,
    \end{equation}
    where $R_{\text{bw}}=100$ kbps and $B_{\text{ref}}=24576$ bits is the 0.167-bpp DeepJSCC/WITT reference payload. These metrics exclude computation latency and energy.
    \item \textbf{Bandwidth efficiency}: PSNR/bpp (dB/bpp).
\end{itemize}

\subsubsection{Comparison Methods}
\label{subsec:baseline}

To comprehensively evaluate the rate-distortion performance and channel robustness of the proposed method, six representative comparison baselines spanning traditional source coding and recent JSCC architectures are selected:

\begin{itemize}
    \item \textbf{BPG (Better Portable Graphics)}: An image compression standard employing HEVC intra-frame coding. Different bitrate points are obtained by adjusting the quantization parameter (QP $\in [44, 51]$), covering the low-bitrate range of bpp $\in [0.06, 0.20]$. As a pure source coding method, BPG's PSNR is measured under noiseless channel conditions as a source-coding reference; under Rayleigh fading, BPG is cascaded with LDPC channel coding as a practical separation baseline.

    \item \textbf{DeepJSCC (Deep Joint Source-Channel Coding)}~\cite{Bourtsoulatze19}: A CNN-based end-to-end analog transmission scheme, in which the encoder directly maps images to complex-valued channel symbols, and the decoder reconstructs images after transmission through the channel. Channel usage rates of $\text{cpp} \in \{1/24, 1/12, 1/6, 1/4\}$ are selected, with equivalent bitrates under QPSK modulation computed as $\text{bpp}_{\text{eq}} = \text{cpp} \times 2 \in \{0.083, 0.167, 0.333, 0.500\}$.

    \item \textbf{WITT (Wireless Image Transmission Transformer)}~\cite{Yang23}: A Transformer-based JSCC scheme that leverages self-attention mechanisms to enhance the expressive power of channel symbols. Channel bandwidth ratios of $\text{CBR} \in \{0.0625, 0.0833\}$, corresponding to $C \in \{96, 128\}$, are selected, with equivalent bitrates under QPSK modulation computed as $\text{bpp}_{\text{eq}} = \text{CBR} \times 2 \in \{0.125, 0.167\}$.

    \item \textbf{SwinJSCC}~\cite{SwinJSCC24}: A Swin Transformer-based JSCC scheme that employs hierarchical self-attention with shifted windows to efficiently capture long-range spatial dependencies in channel symbol mapping.

    \item \textbf{MambaJSCC}~\cite{MambaJSCC24}: A state-space model (Mamba)-based JSCC scheme that leverages linear-time sequence modeling for efficient channel symbol encoding and decoding.

    \item \textbf{DF-JSCC}~\cite{DiffJSCC24}: A diffusion-aided JSCC scheme that leverages generative diffusion priors to enhance perceptual reconstruction quality, providing enhanced robustness under varying channel conditions.
\end{itemize}

To enable comparison of transmission schemes with different coding mechanisms on a unified coordinate system, for analog transmission schemes, this paper uniformly converts their channel usage rates to equivalent bpp based on QPSK modulation (2 bits per channel symbol). This conversion approach is a widely adopted bandwidth normalization method in the semantic communication field. The bpp of the proposed method and BPG is directly computed from the actual number of discrete bits generated, and the proposed bpp includes the 576-bit mask-map overhead. All methods are evaluated under $384 \times 384$ resolution and AWGN channel (SNR = 13 dB baseline) conditions unless otherwise specified. For a fair interpretation, BPG is treated as a source-coding reference, JSCC schemes are treated as channel-aware learned transmission baselines, and the proposed method is evaluated as an edge-assisted semantic transmission scheme whose receiver uses a generative prior. The comparisons are therefore not intended to claim dominance over every possible implementation of each baseline; rather, they quantify the gain of selective token transmission under a consistent resolution, SNR, channel model, and bitrate accounting protocol.

%-------------------------------------------------------------

\subsection{Results and Analysis}
\label{subsec:results}

\begin{figure}[htbp]
    \centering
    \includegraphics[width=0.85\linewidth]{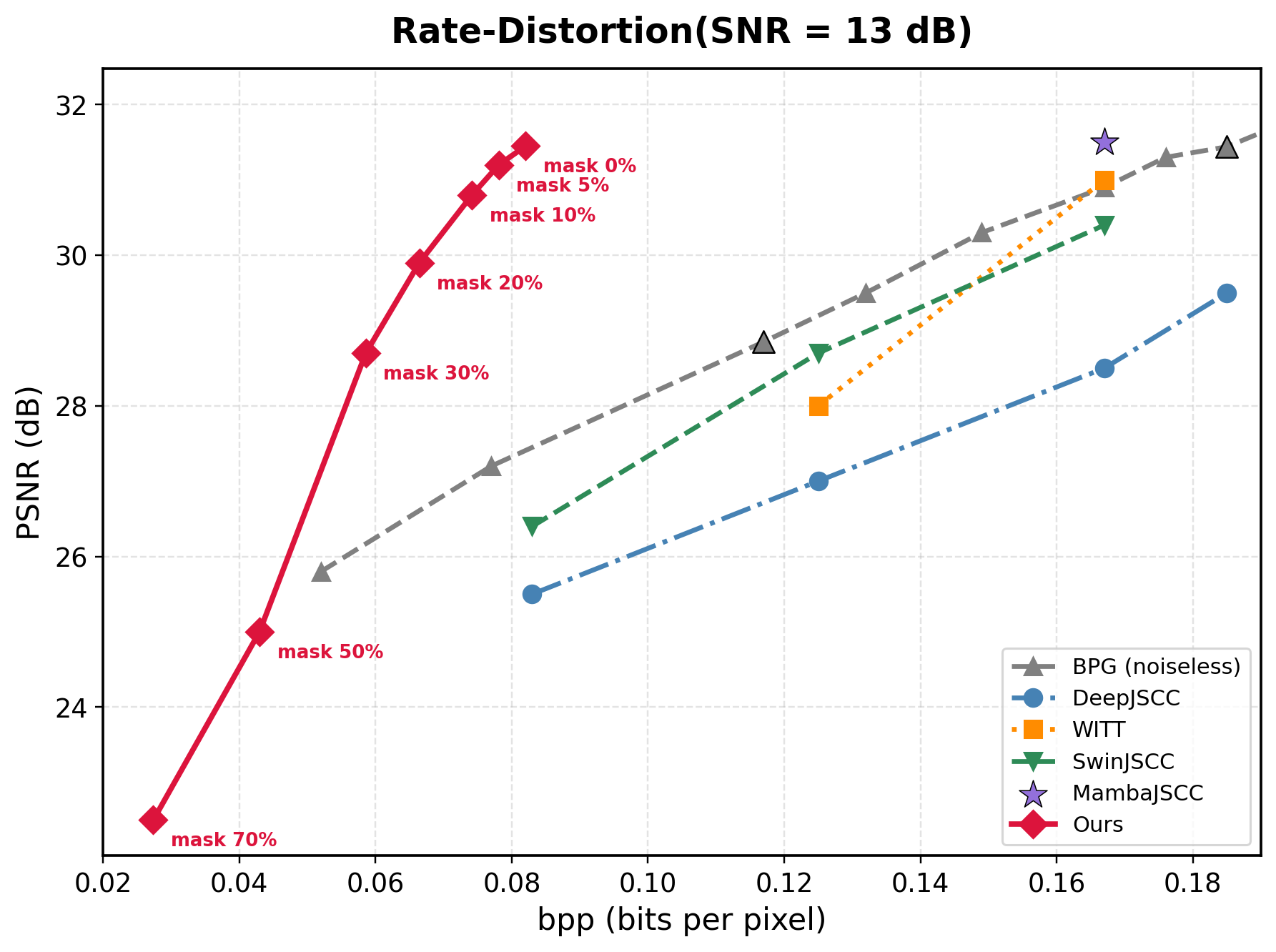}
    \caption{Rate-distortion performance comparison across methods (SNR = 13 dB, AWGN, QPSK).}
    \label{fig:rd_performance}
\end{figure}

\textbf{1) Rate-Distortion Performance.} Fig.~\ref{fig:rd_performance} presents the rate-distortion curves of all methods in the low bitrate region (bpp $\in [0.02, 0.20]$), with bpp on the horizontal axis and PSNR on the vertical axis. All data are measured under SNR = 13 dB, AWGN channel, and QPSK modulation conditions, except that BPG is shown as a noiseless source-coding reference.

The following conclusions can be drawn from the RD curves:

\textbf{a) PSNR advantage at low bitrate.} At bpp $\approx 0.082$, the proposed method (0\% mask) forms a near-equal-bitrate comparison point with DeepJSCC (cpp = 1/24, bpp = 0.083) and SwinJSCC (CBR = 1/24, bpp = 0.083). The proposed method achieves the highest PSNR among the JSCC-based methods at this bitrate in Fig.~\ref{fig:rd_performance}, indicating that the transmitted token anchors and receiver-side generative prior are effective in this low-bitrate setting. BPG is included as a noiseless source-coding reference rather than as a channel-impaired learned transmission method.

\textbf{b) Operation below the tested JSCC bitrate range.} DeepJSCC's minimum channel usage rate in the evaluated configuration (cpp = 1/24) corresponds to bpp = 0.083. The proposed method, by increasing the mask ratio, can reduce the bitrate to 0.027~bpp (70\% mask)---approximately one-third of this tested DeepJSCC point---while still producing recognizable reconstructions. This regime is relevant to narrowband visual IoT links where transmitting a complete learned feature representation may be costly.

\textbf{c) Bandwidth efficiency.} Across the evaluated bpp range, the proposed method obtains a higher PSNR/bpp ratio than the compared baselines in Table~\ref{tab:efficiency}. This reflects the benefit of spending the wireless payload on selected semantic anchors while reconstructing more predictable content at the edge/cloud receiver.

%-------------------------------------------------------------

\textbf{2) Channel Robustness.}

To evaluate the robustness of the proposed method under different channel conditions, this section compares the reconstruction PSNR of each method over the SNR range of $[1, 15]$ dB at a fixed bitrate (bpp $\approx$ 0.082, 0\% mask for the proposed method). Two representative channel models are considered: (1) AWGN channel with QPSK modulation, and (2) Rayleigh fading channel with MMSE equalization.

\textbf{\textit{AWGN Channel.}}

\begin{figure}[htbp]
    \centering
    \includegraphics[width=0.85\linewidth]{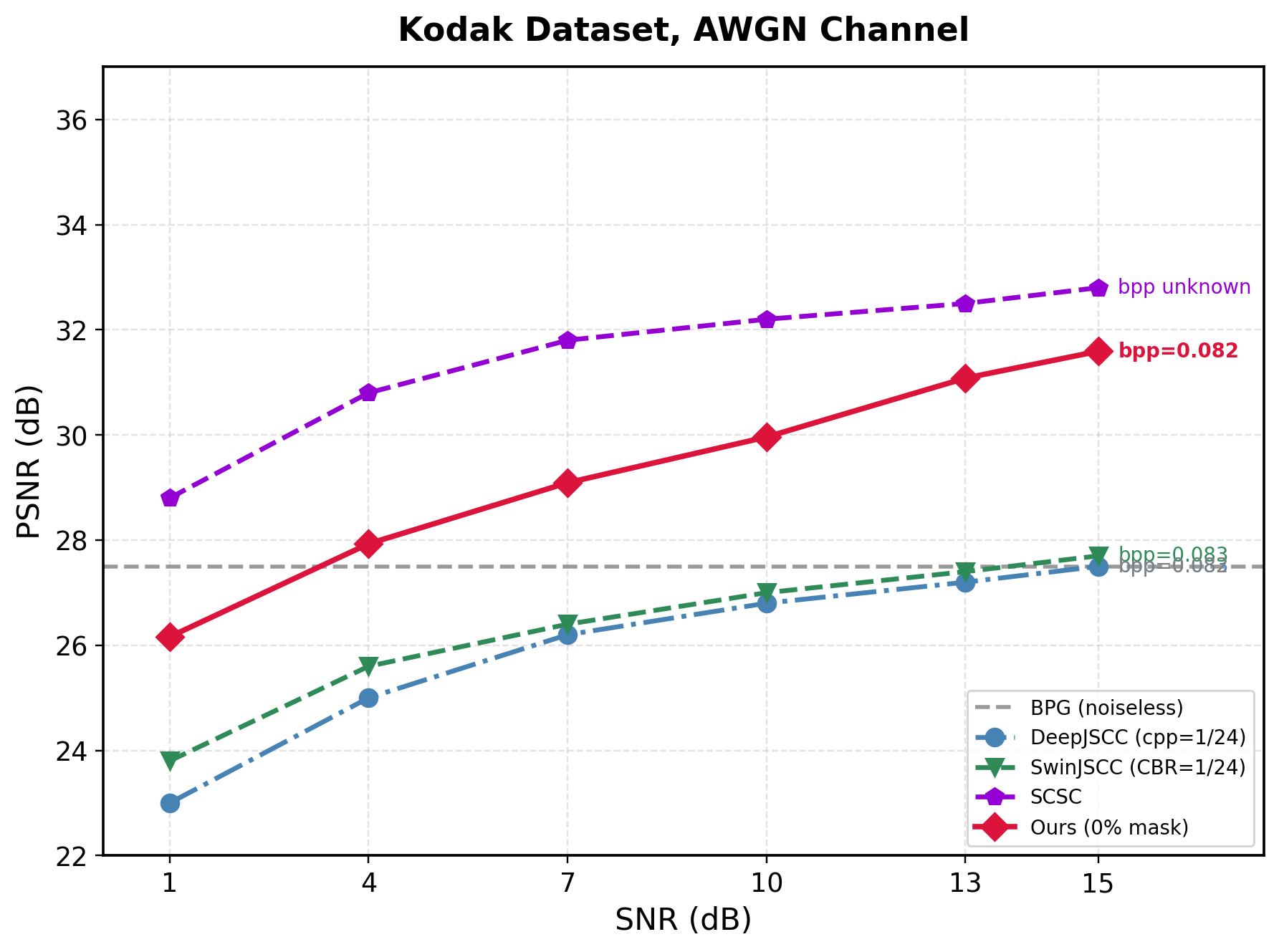}
    \caption{PSNR versus SNR curves under AWGN channel at equal bitrate (bpp $\approx$ 0.082, QPSK).}
    \label{fig:snr_psnr_awgn}
\end{figure}

Fig.~\ref{fig:snr_psnr_awgn} shows the PSNR versus SNR curves under the AWGN channel. BPG, as a noiseless source coding method, has a PSNR that does not vary with SNR and is shown as a dashed reference line. The proposed method achieves higher PSNR than DeepJSCC and SwinJSCC across the tested SNR range. At moderate-to-high SNR, its reconstruction PSNR can exceed the BPG reference at a similar displayed bitrate, which is attributable to the use of a receiver-side generative prior rather than to lossless preservation of all source details.

\textbf{\textit{Rayleigh Fading Channel.}}

\begin{figure}[htbp]
    \centering
    \includegraphics[width=0.85\linewidth]{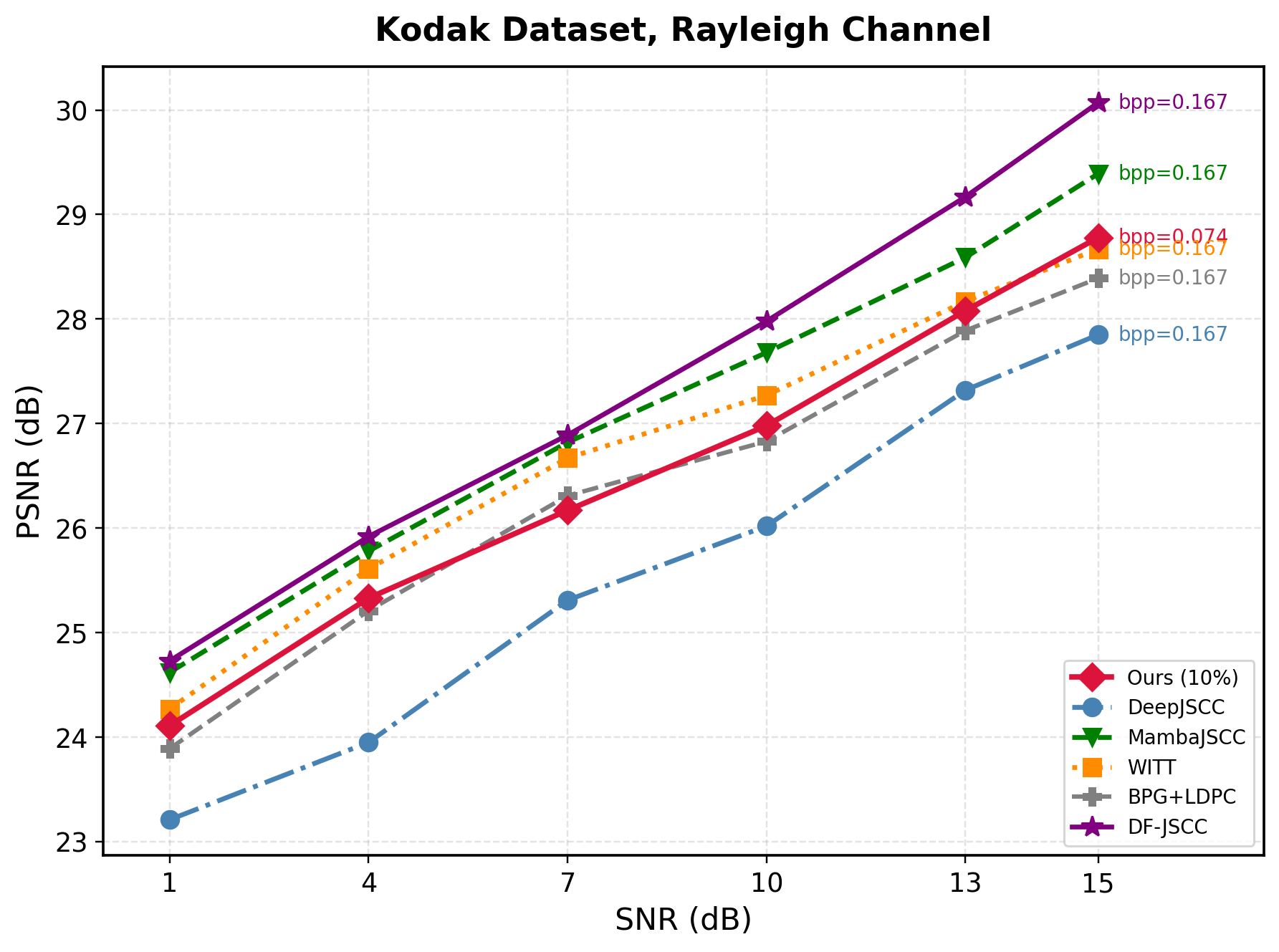}
    \caption{PSNR versus SNR curves under Rayleigh fading channel (QPSK, MMSE equalization). Proposed method at bpp $\approx$ 0.074; comparison methods at bpp = 0.167.}
    \label{fig:snr_psnr_rayleigh}
\end{figure}

Fig.~\ref{fig:snr_psnr_rayleigh} shows the PSNR versus SNR curves under Rayleigh fading with MMSE equalization. The proposed method (10\% mask, bpp = 0.074) operates at a lower bitrate than the compared 0.167-bpp configurations, yet achieves competitive reconstruction quality. At low-to-moderate SNR, the proposed method matches or exceeds several JSCC schemes despite consuming 44.6\% of their link bandwidth. DF-JSCC, benefiting from its diffusion-based generative prior, achieves slightly higher PSNR at moderate-to-high SNR, but at 2.25$\times$ the bitrate of the proposed method. The consistent SNR-PSNR trend across both AWGN and Rayleigh channels indicates that the discrete token transmission and generative recovery framework can be applied to multiple channel models without architectural modification.

From Figures~\ref{fig:snr_psnr_awgn} and~\ref{fig:snr_psnr_rayleigh}, the following conclusions can be drawn:

\textbf{a) PSNR behavior across both channels.} The proposed method demonstrates competitive reconstruction quality under both AWGN and Rayleigh fading channels. Under AWGN, the proposed method obtains higher PSNR than DeepJSCC and SwinJSCC across the tested SNR range. Under Rayleigh fading, the proposed method, operating at a lower bitrate than the compared JSCC configurations, achieves comparable or better PSNR than several JSCC schemes in the low-to-moderate SNR regime.

\textbf{b) Performance preservation at low SNR.} Under severe channel conditions, the proposed method maintains recognizable reconstructions under both channels. The performance gap relative to several comparison methods is preserved at lower SNR values, indicating that token-level selective transmission can improve robustness when the receiver has enough reliable token anchors.

\textbf{c) Cross-channel robustness.} The proposed method's SNR-PSNR trajectory follows a similar trend under both AWGN and Rayleigh channels, with a consistent offset reflecting the additional fading penalty. This demonstrates that the discrete token transmission and generative recovery framework generalizes across channel models without architectural modification.

%-------------------------------------------------------------

\textbf{3) Transmission Efficiency.}

Beyond reconstruction quality, visual IoT deployment also depends on link occupancy and RF-side transmission cost. Table~\ref{tab:efficiency} reports communication-side delay and energy proxy at SNR = 13 dB and QPSK. These values measure only the wireless payload reduction; computation latency and energy are excluded from the table and discussed after it.

\begin{table*}[htbp]
    \centering
    \caption{Communication-side efficiency comparison (SNR = 13 dB, QPSK). Computational latency and GPU energy are excluded.}
    \label{tab:efficiency}
    \footnotesize
    \setlength{\tabcolsep}{3.5pt}
    \begin{tabular}{c c c c c c c}
        \hline
        Method & bpp & $B_{\text{total}}$ (bit) & $T_{\text{comm}}$ (ms) & $E_{\text{comm}}$ (\%) & PSNR (dB) & PSNR/bpp \\
        \hline
        Ours (70\% mask)    & 0.027 & 4,036  & 40.4  & 16.4 & 22.5 & 833.3 \\
        Ours (50\% mask)    & 0.043 & 6,336  & 63.4  & 25.8 & 25.3 & 588.4 \\
        Ours (30\% mask)    & 0.059 & 8,656  & 86.6  & 35.2 & 28.3 & 479.7 \\
        Ours (10\% mask)    & 0.074 & 10,956 & 109.6 & 44.6 & 29.9 & 404.1 \\
        DeepJSCC (cpp=1/12) & 0.167 & 24,576 & 245.8 & 100.0 & 28.8 & 172.5 \\
        WITT (C=128)        & 0.167 & 24,576 & 245.8 & 100.0 & 29.4 & 176.0 \\
        BPG (noiseless)     & 0.167 & 24,576 & ---   & ---   & 31.3 & 187.4 \\
        \hline
    \end{tabular}
\end{table*}

At bpp $\approx$ 0.074 (10\% mask), the proposed method uses 44.6\% of the 0.167-bpp DeepJSCC/WITT payload while achieving higher PSNR in this experiment. The 70\% mask configuration further reduces the RF-energy proxy to 16.4\% of the same reference. These values reflect communication-side savings only; Detic analysis and 32-step MaskGIT reconstruction introduce separate inference cost, making the method most suitable for gateway- or edge-server-assisted visual IoT links.

\textbf{4) Ablation Study on Mask Generation.}

We evaluate the semantic-aware mask at SNR = 13 dB without changing bitrate. With all tokens retained, VQ reconstruction reaches 32.47 dB PSNR, 0.951 SSIM, and 0.075 LPIPS. At 30\% mask (0.059 bpp), the proposed mask achieves 28.30 dB / 0.900 / 0.118, outperforming random masking by 0.85 dB / 0.015 / 0.009. At 50\% mask (0.043 bpp), the gain increases to 1.30 dB / 0.017 / 0.014, showing that semantic selection becomes more important as the link budget decreases.

At 30\% mask, the full mask generator achieves 28.30 dB PSNR, 0.900 SSIM, and 0.118 LPIPS, improving over Editability Only by 0.73 dB / 0.011 / 0.008 and over Semantic Only by 0.92 dB / 0.010 / 0.006. This confirms that recovery difficulty and semantic importance provide complementary cues.

\textbf{5) Downstream Object Detection.}

Pixel-level metrics do not directly measure whether task-relevant objects survive the communication pipeline, so we evaluate object detection as a downstream semantic proxy.

\textbf{Setup.} Mask R-CNN with ResNet-50-FPN, pretrained on MS-COCO, is used as an independent evaluator distinct from Detic. For each Kodak image, Mask R-CNN detections on the original image are used as pseudo-GT; detections on reconstructed images are then evaluated at mAP@0.5 with confidence threshold 0.3 under 30\% and 50\% masks at SNR = 13 dB.

\textbf{Results.} Table~\ref{tab:downstream_map} reports mAP@0.5 for both masking strategies.

\begin{table}[htbp]
\centering
\footnotesize
\setlength{\tabcolsep}{3.5pt}
\caption{Downstream object detection mAP@0.5 on reconstructed images. Pseudo-GT denotes reference detections obtained from the original (uncompressed) image.}
\label{tab:downstream_map}
\begin{tabular}{ccc}
\hline
\textbf{Ratio} & \textbf{Type} & \textbf{mAP@0.5}  \\
\hline
% & Original (GT) & 0.7813 & 8.40 \\
30\% & Random        & 0.5856  \\
30\% & Semantic      & \textbf{0.6742} \\
\hline
50\% & Random        & 0.3962 \\
50\% & Semantic      & \textbf{0.4142} \\
\hline
\end{tabular}
\end{table}

At identical bitrate, the semantic-aware mask improves mAP@0.5 over random masking at both mask ratios, with a larger gain at 30\% masking. This supports downstream machine perception as a useful evaluation dimension beyond pixel fidelity.

\textbf{6) Cross-Domain Evaluation on VisDrone IoT Scenes.}

We further evaluate cross-domain reconstruction on the VisDrone2019-DET test-dev dataset, which contains 1,000 drone-captured aerial images. This experiment focuses on reconstruction robustness rather than annotated detection mAP.

\begin{figure}[htbp]
    \centering
    \includegraphics[width=0.85\linewidth]{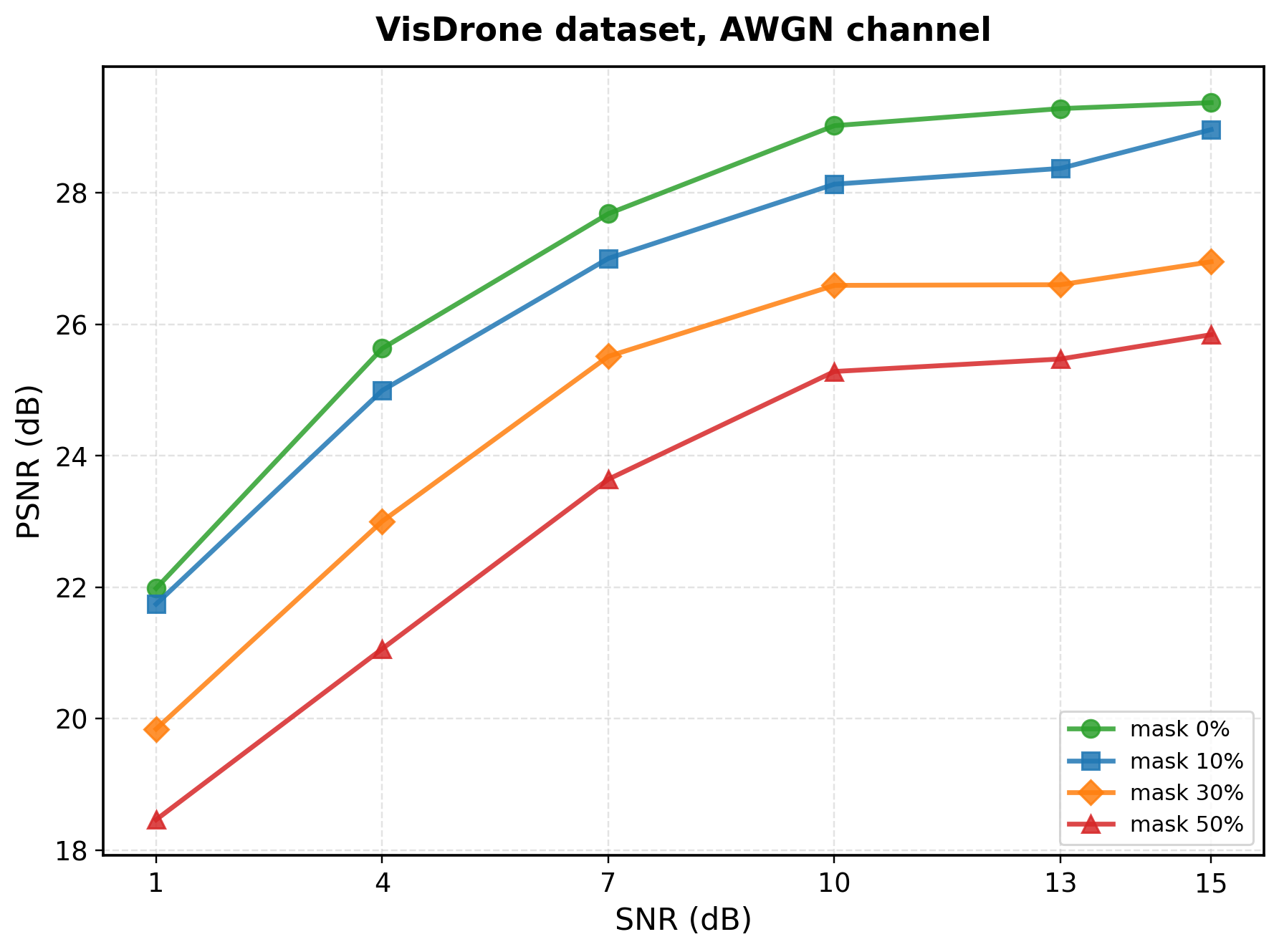}
    \caption{PSNR versus SNR curves on the VisDrone2019 dataset under AWGN channel at different mask ratios.}
    \label{fig:visdrone_snr}
\end{figure}

Fig.~\ref{fig:visdrone_snr} shows AWGN results for 0\%, 10\%, 30\%, and 50\% masks over $\{1,4,7,10,13,15\}$ dB. PSNR increases with SNR for all mask ratios, and lower mask ratios retain higher reconstruction quality.

\begin{figure}[htbp]
    \centering
    \includegraphics[width=0.85\linewidth]{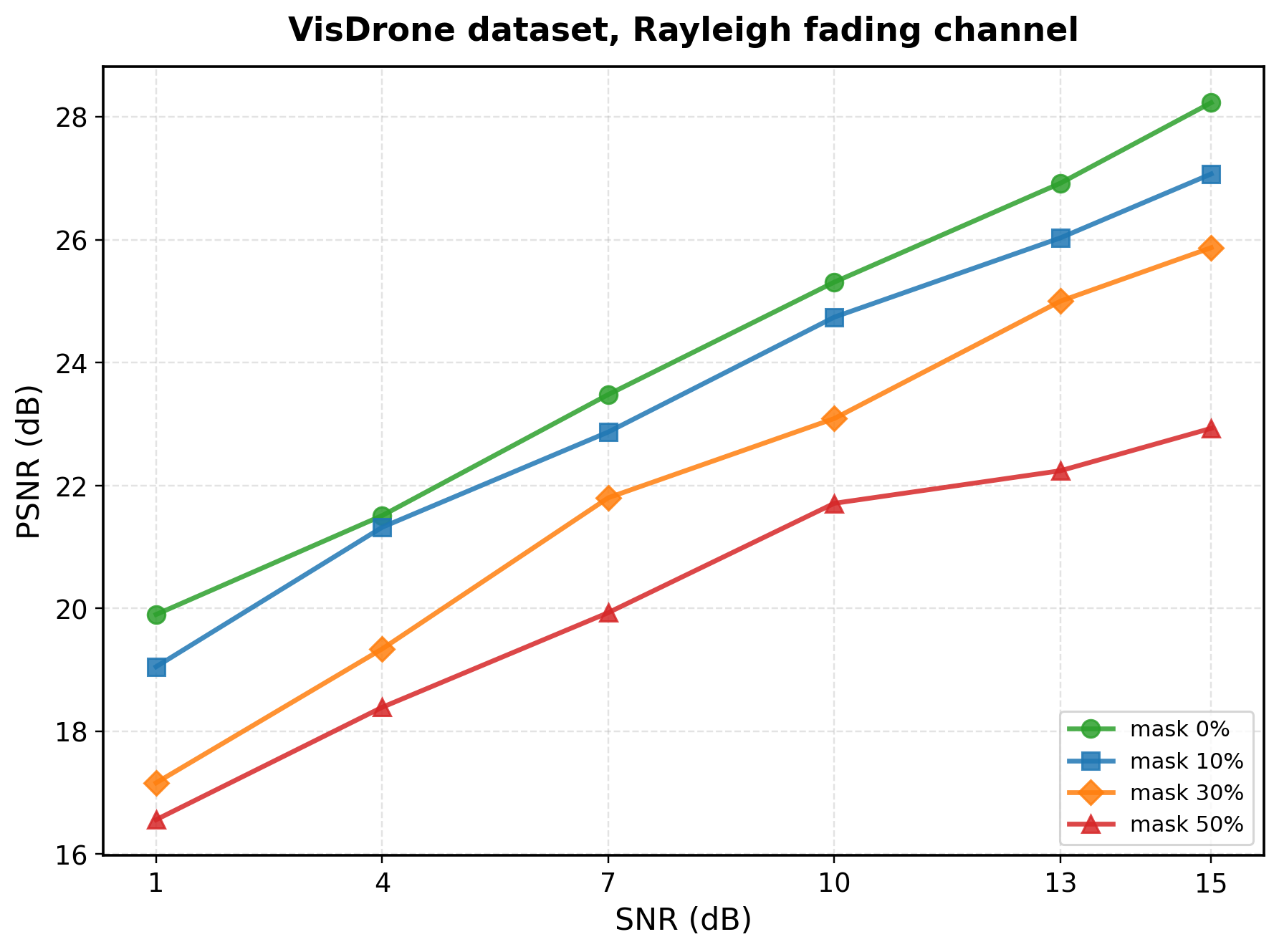}
    \caption{PSNR versus SNR curves on the VisDrone2019 dataset under Rayleigh fading channel (MMSE equalization) at different mask ratios.}
    \label{fig:visdrone_rayleigh}
\end{figure}

Fig.~\ref{fig:visdrone_rayleigh} reports the corresponding Rayleigh-fading results with MMSE equalization. Although fading causes a PSNR penalty relative to AWGN, the SNR trend and ordering among mask ratios remain consistent, indicating stable reconstruction behavior without VisDrone-specific fine-tuning.

\textbf{7) Practical Deployment and Scope of Claims.}

The evaluation isolates the communication benefit of semantic token selection under an edge-assisted visual IoT architecture. Thus, delay and RF-energy values quantify wireless payload reduction rather than total device energy. VQ encoding, semantic analysis, and MaskGIT reconstruction may be placed on the camera, gateway, or cloud depending on hardware capability; the most suitable setting is gateway- or edge-server-assisted visual IoT. Since Kodak has no object annotations, pseudo-GT detections are used only as a semantic consistency reference, and the VisDrone experiment does not claim annotated detection mAP.

\section{Conclusion}
\label{sec:conclusion}

This paper investigates, from the perspective of resource-constrained visual IoT, how to select and organize visual semantic information for transmission over constrained wireless links. Unlike existing work that primarily focuses on encoder architecture design or channel modeling, this paper addresses the explicit modeling and scheduling of semantic information at the token-level spatial dimension, and proposes a semantic-aware mask generation method based on MaskGIT along with an edge-assisted generative image transmission framework.

The proposed method performs joint modeling through two complementary dimensions---base editability estimation and semantic importance protection---and generates a semantic-aware mask via score fusion and spatial dispersal sampling, enabling the IoT transmitter or gateway to protect critical semantic regions and hard-to-recover structures while leaving relatively redundant and recoverable content to be completed by the edge/cloud-side MaskGIT model. The IoT transmitter sends only the quantization indices of kept tokens and a binary mask map, and the edge/cloud receiver employs parallel iterative sampling with Halton sequence scheduling to recover missing tokens.

Experimental results demonstrate that: (1) under the evaluated equal-bitrate settings, the proposed method achieves higher reconstruction quality than the tested DeepJSCC and WITT configurations, with improved PSNR/bpp efficiency; (2) by increasing the mask ratio, the method can flexibly trade reconstruction quality for reductions in communication-link delay and RF transmission energy, although neural inference cost remains a separate deployment consideration; (3) the semantic-aware mask strategy outperforms random masking in the tested reconstruction and pseudo-GT downstream detection settings, validating the effectiveness of token-level semantic selection within the evaluated protocol. Taken together, the proposed method achieves a favorable balance between communication efficiency and semantic fidelity through the collaborative mechanism of ``intelligent selection at the IoT side, generative restoration at the edge/cloud side.''

Future work will further explore temporally consistent mask modeling for visual IoT video streams, end-to-end optimization incorporating downstream visual task accuracy (e.g., object detection mAP, semantic segmentation mIoU), lightweight IoT-side semantic extraction, annotated task evaluation on larger IoT datasets, and more efficient channel coding schemes to further reduce the actual IoT link overhead.

\bibliographystyle{IEEEtran}
\bibliography{refs}

@article{Shannon48,
  author  = {C. E. Shannon},
  title   = {A mathematical theory of communication},
  journal = {Bell Syst. Tech. J.},
  volume  = {27},
  number  = {3},
  pages   = {379--423},
  year    = {1948},
}

@article{Halton60,
  author  = {J. H. Halton},
  title   = {On the efficiency of certain quasi-random sequences of points in evaluating multi-dimensional integrals},
  journal = {Numer. Math.},
  volume  = {2},
  number  = {1},
  pages   = {84--90},
  year    = {1960},
}

@inproceedings{Vaswani17,
  author    = {A. Vaswani and N. Shazeer and N. Parmar and J. Uszkoreit and L. Jones and A. N. Gomez and L. Kaiser and I. Polosukhin},
  title     = {Attention is all you need},
  booktitle = {Proc. Adv. Neural Inf. Process. Syst. (NeurIPS)},
  pages     = {5998--6008},
  year      = {2017},
}

@inproceedings{Chang22,
  author    = {H. Chang and H. Zhang and L. Jiang and C. Liu and W. T. Freeman},
  title     = {{MaskGIT}: Masked generative image transformer},
  booktitle = {Proc. IEEE/CVF Conf. Comput. Vis. Pattern Recognit. (CVPR)},
  pages     = {11315--11325},
  year      = {2022},
}

@inproceedings{VanDenOord17,
  author    = {A. van den Oord and O. Vinyals and K. Kavukcuoglu},
  title     = {Neural discrete representation learning},
  booktitle = {Proc. Adv. Neural Inf. Process. Syst. (NeurIPS)},
  pages     = {6306--6315},
  year      = {2017},
}

@article{Getu24,
  author  = {T. M. Getu and G. Kaddoum and M. Bennis},
  title   = {Semantic communication: A survey on research landscape, challenges, and future directions},
  journal = {Proc. IEEE},
  volume  = {112},
  number  = {11},
  pages   = {1--30},
  year    = {2024},
}

@article{Nguyen25,
  author  = {L. X. Nguyen and A. D. Raha and P. S. Aung and D. Niyato and Z. Han and C. S. Hong},
  title   = {A contemporary survey on semantic communications: Theory of mind, generative {AI}, and deep joint source-channel coding},
  journal = {IEEE Commun. Surveys Tuts.},
  year    = {2025},
}

@article{Liang24,
  author  = {C. Liang and H. Du and Y. Sun and D. Niyato and J. Kang and D. Zhao and M. A. Imran},
  title   = {Generative {AI}-driven semantic communication networks: Architecture, technologies and applications},
  journal = {IEEE Trans. Cogn. Commun. Netw.},
  volume  = {10},
  number  = {5},
  pages   = {1911--1931},
  year    = {2024},
}

@article{Huynh24,
  author  = {N. V. Huynh and J. Wang and H. Du and D. T. Hoang and D. Niyato and D. N. Nguyen and D. I. Kim and K. B. Letaief},
  title   = {Generative {AI} for physical layer communications: A survey},
  journal = {IEEE Trans. Cogn. Commun. Netw.},
  volume  = {10},
  number  = {3},
  pages   = {706--728},
  year    = {2024},
}

@article{Huang24,
  author  = {C.-H. Huang and J.-L. Wu},
  title   = {Unveiling the future of human and machine coding: A survey of end-to-end learned image compression},
  journal = {Entropy},
  volume  = {26},
  number  = {5},
  pages   = {357},
  year    = {2024},
}

@article{Li25TokenSurvey,
  author  = {J. Li and X. Wang and Y. Zhang and others},
  title   = {Discrete visual tokenization: A comprehensive survey of vector quantization for image generation},
  journal = {arXiv preprint arXiv:2504.14807},
  year    = {2025},
}

@article{SwinJSCC24,
  author  = {K. Yang and S. Wang and J. Dai and X. Qin and K. Niu and P. Zhang},
  title   = {{SwinJSCC}: Taming {Swin} Transformer for deep joint source-channel coding},
  journal = {IEEE Trans. Cogn. Commun. Netw.},
  volume  = {11},
  number  = {1},
  pages   = {90--104},
  year    = {2024},
}

@inproceedings{MambaJSCC24,
  author    = {T. Wu and Z. Chen and M. Tao and Y. Sun and X. Xu and W. Zhang and P. Zhang},
  title     = {{MambaJSCC}: Adaptive deep joint source-channel coding with generalized state space model},
  booktitle = {Proc. IEEE Global Commun. Conf. (GLOBECOM)},
  year      = {2024},
}

@inproceedings{DiffJSCC24,
  author    = {S. F. Yilmaz and C. Karakus and D. G\"{u}nd\"{u}z},
  title     = {Diffusion-aided joint source channel coding for high realism wireless image transmission},
  booktitle = {Proc. IEEE Int. Conf. Commun. (ICC)},
  year      = {2024},
}

@article{Gunduz24,
  author  = {D. G\"{u}nd\"{u}z and M. A. Wigger and T. M. Getu and others},
  title   = {Joint source-channel coding: Fundamentals and recent progress in practical designs},
  journal = {arXiv preprint arXiv:2409.17557},
  year    = {2024},
}

@article{Bian24,
  author  = {C. Bian and Y. Shao and H. Wu and E. Ozfatura and D. G\"{u}nd\"{u}z},
  title   = {Process-and-forward: Deep joint source-channel coding over cooperative relay networks},
  journal = {IEEE J. Sel. Areas Commun.},
  year    = {2024},
}

@article{MNTSCC25,
  author  = {C. Wang and C. Li and Y. Liao and C. Ding and Z. Ye},
  title   = {{MNTSCC}: A {VMamba}-based nonlinear joint source-channel coding for semantic communications},
  journal = {Comput. Mater. Continua},
  volume  = {85},
  number  = {2},
  year    = {2025},
}

@inproceedings{SING25,
  author    = {J. Chen and S. F. Yilmaz and D. You and P. L. Dragotti and D. G\"{u}nd\"{u}z},
  title     = {{SING}: Semantic image communications using null-space and {INN}-guided diffusion models},
  booktitle = {Proc. IEEE Int. Conf. Commun. (ICC)},
  year      = {2025},
}

@inproceedings{Besnier25,
  author    = {V. Besnier and M. Chen and D. Hurych and E. Valle and M. Cord},
  title     = {Halton scheduler for masked generative image transformer},
  booktitle = {Proc. Int. Conf. Learn. Represent. (ICLR)},
  year      = {2025},
}

@inproceedings{MaskBit25,
  author    = {M. Weber and L. Yu and Q. Yu and X. Deng and X. Shen and D. Cremers and L.-C. Chen},
  title     = {{MaskBit}: Embedding-free image generation via bit tokens},
  booktitle = {Proc. Int. Conf. Learn. Represent. (ICLR)},
  year      = {2025},
}

@inproceedings{TokenFlow25,
  author    = {L. Qu and S. Liu and H. Zhang and X. Chen and X. Wang and Y. Jiang},
  title     = {{TokenFlow}: Unified image tokenizer for multimodal understanding and generation},
  booktitle = {Proc. IEEE/CVF Conf. Comput. Vis. Pattern Recognit. (CVPR)},
  year      = {2025},
}

@article{UniTok25,
  author  = {M. Chen and S. Liu and J. Wang and others},
  title   = {{UniTok}: A unified tokenizer for visual generation and understanding},
  journal = {arXiv preprint arXiv:2502.20321},
  year    = {2025},
}

@inproceedings{Esser21,
  author    = {P. Esser and R. Rombach and B. Ommer},
  title     = {Taming transformers for high-resolution image synthesis},
  booktitle = {Proc. IEEE/CVF Conf. Comput. Vis. Pattern Recognit. (CVPR)},
  pages     = {12873--12883},
  year      = {2021},
}

@article{Bourtsoulatze19,
  author  = {E. Bourtsoulatze and D. Burth Kurka and D. G\"{u}nd\"{u}z},
  title   = {Deep joint source-channel coding for wireless image transmission},
  journal = {IEEE Trans. Cogn. Commun. Netw.},
  volume  = {5},
  number  = {3},
  pages   = {567--579},
  year    = {2019},
}

@inproceedings{Yang23,
  author    = {M. Yang and C. Bian and H.-S. Kim},
  title     = {{WITT}: A wireless image transmission transformer for semantic communications},
  booktitle = {Proc. IEEE Int. Conf. Acoust., Speech Signal Process. (ICASSP)},
  pages     = {1--5},
  year      = {2023},
}

@article{Zhao25,
  author  = {C. Zhao and H. Du and D. Niyato and J. Kang and Z. Xiong and D. I. Kim and X. Shen and K. B. Letaief},
  title   = {Generative {AI} for secure physical layer communications: A survey},
  journal = {IEEE Trans. Cogn. Commun. Netw.},
  volume  = {11},
  number  = {1},
  pages   = {3--26},
  year    = {2025},
}

@article{Wang25,
  author  = {Y. Wang and H. Han and Y. Feng and J. Zheng and B. Zhang},
  title   = {Semantic communication empowered {6G} networks: Techniques, applications, and challenges},
  journal = {IEEE Access},
  volume  = {13},
  year    = {2025},
}

@article{Zhong25,
  author  = {R. Zhong and X. Mu and M. Jaber and Y. Liu},
  title   = {Enabling distributed generative {AI} in {6G}: Mobile edge generation},
  journal = {IEEE Internet Things J.},
  volume  = {12},
  number  = {6},
  pages   = {6607--6620},
  year    = {2025},
}

@inproceedings{Yuan25,
  author  = {X. Yuan and J. Ren and Y. Wang and Z. Wang and X. Feng and H. Kim and C. Wu},
  title   = {Generative semantic communication for joint image transmission and segmentation},
  booktitle = {Proc. IEEE Int. Conf. Commun. (ICC)},
  year    = {2025},
}

@article{HernandezCamara25,
  author  = {P. Hern\'{a}ndez-C\'{a}mara and J. Vila-Tom\'{a}s and V. Laparra and J. Malo},
  title   = {Dissecting the effectiveness of deep features as metric of perceptual image quality},
  journal = {Neural Netw.},
  volume  = {185},
  pages   = {107189},
  year    = {2025},
}

\end{document}